# Genetic Algorithm Based Combinatorial Optimization for the Optimal Design of Water Distribution Network of Gurudeniya Service Zone, Sri Lanka

**K.H.M.R.N. Senavirathna and C.K. Walgampaya**

**Abstract:** This paper brings an in-detail Genetic Algorithm (GA) based combinatorial optimization method used for the optimal design of the water distribution network (WDN) of Gurudeniya Service Zone, Sri Lanka. Genetic Algorithm (GA) mimics the survival of the fittest principle of nature to develop a search process. Methodology employs fuzzy combinations of pipe diameters to check their suitability to be considered as the cost-effective optimal design solutions. Furthermore, the hydraulic constraints were implicitly evaluated within the GA itself in its aim for reaching the global optimum solution. Upon analysis, the results of GA approach delivered satisfactory design outputs.

## 1. Introduction

WDN is a complex infrastructure that involves high costs for implementation. WDN is typically composed of pipes, pumps, valves, consumer outlets or nodes, and storage reservoirs. Of all these components of a WDN, the interconnecting pipes through which water is transported from source node to the demand node account for the major fraction of the capital cost. Hence, it is important to obtain best network layout and the combination of pipe diameters that minimize the implementation cost while satisfying all the hydraulic constraints. As the joint consideration of the network layout and the design is highly complex, and since the pipe layout is almost all the time restricted by the location of the roads, most of the model formulations and solution methods have been proposed and tested only for least-cost design only (Savic & Walters [10]; Alporevits & Shamir [1]; Lin et al. [7]; Cunha & Sousa [4]; Dijk et al. [5]; Mohan & Babu [9]; Suribabu [12]; Cong & Zhao [13]; Awe et al. [2]; Senavirathna et al. [11]).

## 2. Literature Review

Corte & Sörensen [3], in their critical review, have shown that not only closed-conduit hydraulic modeling, but also optimization of WDNs is much needed to reach the goals of implementing a reliable water supply network at optimum cost. In literature, there can be seen two main categories for WDN optimization algorithms, such as, deterministic-heuristic and stochastic-heuristic algorithms.

Deterministic-heuristic algorithms normally use gradient and implicit information associated with the WDN model formulation to reach an optimal solution. The remarkable advantage of algorithms of this kind is that, it takes a very smaller number of iterations to reach the optimal solution compared to stochastic-heuristic algorithms. Moreover, deterministic-heuristic algorithms cannot guarantee if the solution obtained is the global optimum solution, as the solution obtained by them may be some local optimal. The studies done by Lin et al. [7], Hsu & Cheng [6], Mohan & Babu [8], Suribabu [12], and Awe et al. [2], provide criteria for putting deterministic-heuristic algorithms to use in order for the algorithm to reach near optimal solutions for WDNs.

On the other hand, stochastic-heuristic, or sometimes called modern-stochastic-heuristic, methods neither take gradient nor implicit information associated with the WDNs as a norm.

*Eng. **K.H.M.R.N. Senavirathna**, B. Sc. Eng (Hons) (Peradeniya), Temporary Instructor, Department of Engineering Mathematics, Faculty of Engineering, University of Peradeniya.*
*Email:rajithas@eng.pdn.ac.lk*
*ORCID ID: https://orcid.org/0000-0002-4081-3781*

*Eng. **(Dr.) C.K. Walgampaya**, AMIE (Sri Lanka), B.Sc. Eng. (Hons) (Peradeniya), M.Sc.(UofL,KY, USA), Ph.D. (UofL, KY, USA), Senior Lecturer, Department of Engineering Mathematics, Faculty of Engineering, University of Peradeniya.*
*Email:ckw@pdn.ac.lk*
*ORCID ID: https://orcid.org/0000-0003-2192-474X*

These stochastic-heuristic algorithms evaluate the objective function at randomly taken different regions of the solution space to probe the feasible solutions that minimize the cost function. The main advantage of this method is that it investigates the whole solution space to seek the global optimum solution. On the contrary to deterministic-heuristic methods, stochastic-heuristic methods compute very large number of objective function evaluations to reach this global optimum. Some of the notable studies that utilize these stochastic-heuristic optimization algorithms for WDNs include Simulated Annealing (SA) Approach done by Cunha & Sousa [4], Honey-Bee Mating Optimization (HBMO) by Mohan & Babu [9], and Genetic Algorithm (GA) by Savic & Walters [10], and Dijk et al. [5].

This paper, however, brings about a study on the WDN of Gurudeniya Service Zone, Sri Lanka by utilizing the GA based combinatorial optimization for its optimal design process.

## 3. Methodology and Analysis

### 3.1 Genetic Algorithm (GA)

GA mimics the survival of the fittest principle of nature to develop a search process. The original version of the GA can be found in the work of Holland (1975). GA is naturally suitable for maximization problems. For the ease of understanding the working principles of GA, for example, a scenario where $f(x_1, x_2)$ has to be maximized can be considered. An individual (he/she) is represented by point $A(x_1^0, x_2^0)$, and the location $(x_1^0, x_2^0)$ represents the genetic code of that person. The function value $f(x_1^0, x_2^0)$ can then represent the fitness of this person. If $B(x_1^1, x_2^1)$ is another person who is of the opposite sex than $A$, then they can produce children, and consequently generations from $A$ and $B$, imparting their genetic code to the children. If their following generations are to survive, by the natural selection process, they should move towards point $M$ where the genetic code has the maximum fitness, $f(x_1^m, x_2^m)$. In GA, a large population with different fitness is finally moved towards highest fitness by mimicking the natural selection process through three GA operators – *Reproduction*, *Crossover* and *Mutation* - as illustrated by the Figure 1. In the following sections, it is presented the adaptation of GA to tackle the optimal design problem of the water distribution networks (WDNs).

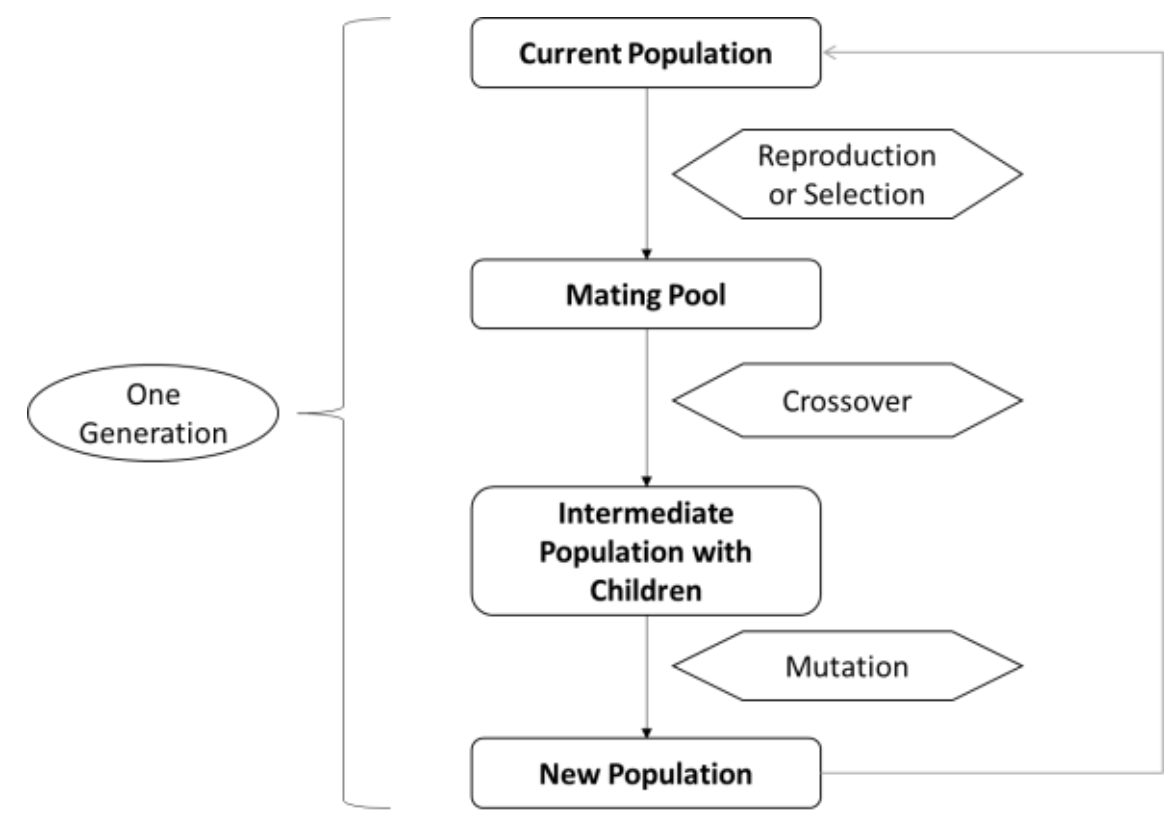

**Figure 1 - Three operators in GA: Reproduction, Crossover and Mutation**

### 3.2 Model Formulation for GA based Combinatorial Optimization of WDN

An example tree-shaped eight-variable WDN is shown in Figure 2 for the ease of understanding the objective function given in equation (1).

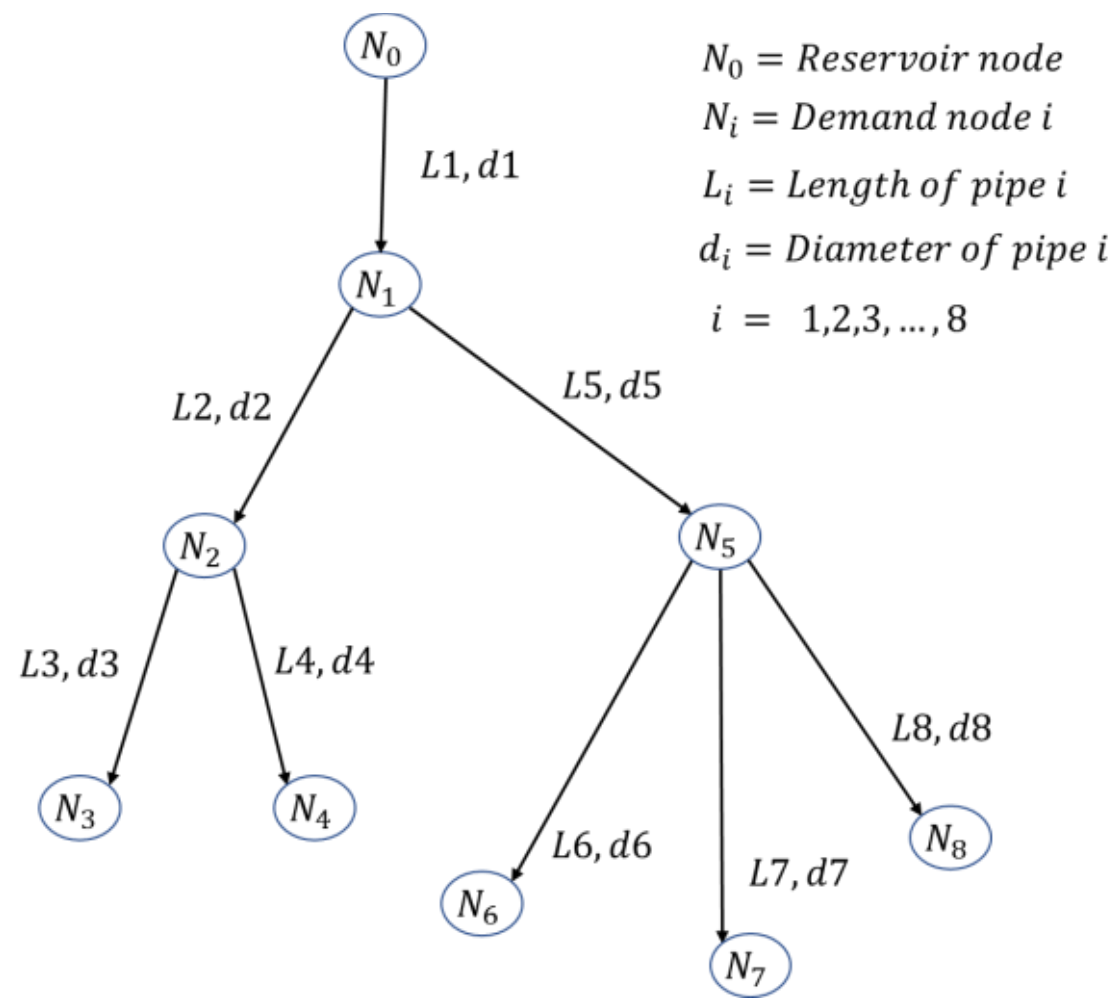

**Figure 2 - An example tree-shaped eight-variable WDN**

The objective function of the WDN design criterion focused in this study can be written as given in equation (1),

$$Min\ Z = \sum_{i=1}^{N} C_i(d, L) \tag{1}$$

subjects to the hydraulic design constraints given in the inequalities (2) and (3),

$$H_{R\ j} \geq H_{R\ j}^{min} \quad j = 1,2,3, \ldots, nd \tag{2}$$

$$g_{FF\ i} \leq g_{FF\ i}^{max} \quad i = 1,2,3, \ldots, np \tag{3}$$

where,

$Z$ =total cost involved with the WDN;

$N$ =number of pipes in the water distribution network;

$C_i(d, L)$ =cost of the $i^{th}$ pipe having diameter $d$ and length $L$;
$H_{R\ j}$ =residual water head available at the $j^{th}$ node;
$H_{R\ j}^{min}$ =minimum residual water head required at the $j^{th}$ node;
$g_{FF\ i}$ =friction-and-fitting-loss-gradient in the $i^{th}$ pipe;
$g_{FF\ i}^{max}$ =maximum allowable friction and fitting loss gradient in the $i^{th}$ pipe;
$nd$ =number of demand nodes; and
$np$ =number of pipes.

The friction-and-fitting-loss-gradient in a pipe can be calculated as equation (4).

$$g_{FF\ i} = C_{ft} \frac{10.666 * Q_i^{1.85}}{C_{HW}^{1.85} * d_i^{4.87}} \quad i = 1,2,3, \ldots, np \tag{4}$$

where,
$C_{ft}$ = fitting loss coefficient;
$C_{HW}$ = Hazen William's Coefficient;
$Q_i$ = water flow in the $i^{th}$ pipe; and
$d_i$ = diameter of the $i^{th}$ pipe.

The continuity constraint at each node is as equation (5).

$$Q_j^{in} - Q_j^{out} - D_j = 0 \quad j = 1,2,3, \ldots, nd \tag{5}$$

where,
$Q_j^{in}$ = flow entering the $j^{th}$ node;
$Q_j^{out}$ = flow leaving the $j^{th}$ node; and
$D_j$ = demand at the $j^{th}$ node.

The mathematical representation of the hydraulic energy balance through a closed loop can be given as equation (6).

$$\left( \sum_{i=1}^{np_L} HL_i \right)_L = 0 \qquad L = 1,2,3, \ldots, nL \tag{6}$$

where,
$HL_i$ = head-loss in pipe $i$;
$np_L$ = number of pipes in a loop; and
$nL$ = number of loops in the WDN.

The adaptation of the GA for the optimal design of the WDN, is shown in the flowchart of the Figure 3. For a typical design, it is initially required the WDN data, such as reservoir elevation, nodal demands, pipe lengths, nodal elevations, and the pipe connectivity layout. In addition to WDN data, commercially available pipe diameters and their unit costs are needed as the fulfillment of the data requirement.

At the beginning, all the GA parameters are defined. These parameters are the coding of the problem parameters - i. e. string length $l$, population size $n$, crossover probability $p_c$, mutation probability $p_m$, and the allowable maximum number of iterations (generations) $t_{max}$.

Once the above GA parameters are assigned, random $n$ number of solutions are generated to be used as the initial population. The fitness $F$ of each solution is then calculated. Subsequently, the creation of the mating pool by performing the reproduction operator, the creation of the intermediate population by performing the crossover operator, and the creation of the new population by performing the mutation operator, are carried out to finally reach the evaluation of the fitness $F$ of the strings of the new population. The solution(s) with the highest fitness is(are) considered as the best of that particular iteration or generation. However, as shown in the Figure 3, many iterations or generations can be performed, until the stopping criterion $t > t_{max}$ (or other) is met, in order to witness to global best solution. Now this global best solution can be considered as the combination of pipe diameters that minimizes the cost of the WDN while satisfying the design constrains to the maximum extent possible.

In the following sections of the methodology, the adaptation of the coding of the problem parameters, the fitness function, and the three GA operators – reproduction, crossover and mutation, are written explicitly for a clear understanding.

### 3.3 Coding

The genetic code of an individual is taken as the list of decision variables involved with the individual. For example, the genetic code of an eight variable WDN problem can be represented as follows:

| F1 | F2 | F3 | F4 | F5 | F6 | F7 | F8 |
|---|---|---|---|---|---|---|---|

where $F1$ to $F8$ are the genes of an individual (F), preferably considered to be a "father" in the population. These gene values are chosen from the list of commercially available pipe diameters, with a degree of belongingness, also known as the membership function, equal to $1/8$ for each gene, for an eight-variable problem.

Start

WDN Data: Reservoir Elevation, Nodal Demands, Pipe Lengths, Nodal Elevations, Connectivity Layout;
Commercially available pipe diameters and their unit costs

Select
- Coding for the problem parameters (string length $l$)
- Population size $n$
- Crossover probability $p_c$
- Mutation probability $p_m$
- The allowable maximum number of iterations (generations) $t_{max}$

Generation of initial random $n$ number of solutions from commercially available pipe diameters, with string length $l$

Set $t = 0$

Evaluate the fitness, $F$, for each string in the population

Is the stopping criterion $t > t_{max}$ (or other) met?

Yes

No

Perform reproduction of the population and create the mating pool

Perform crossover on the random pairs of strings [Now this is the intermediate population]

Perform mutation on every string [Now this is the new population]

Evaluate the fitness, $F$, for the strings in the new population

Set $t = t + 1$

End

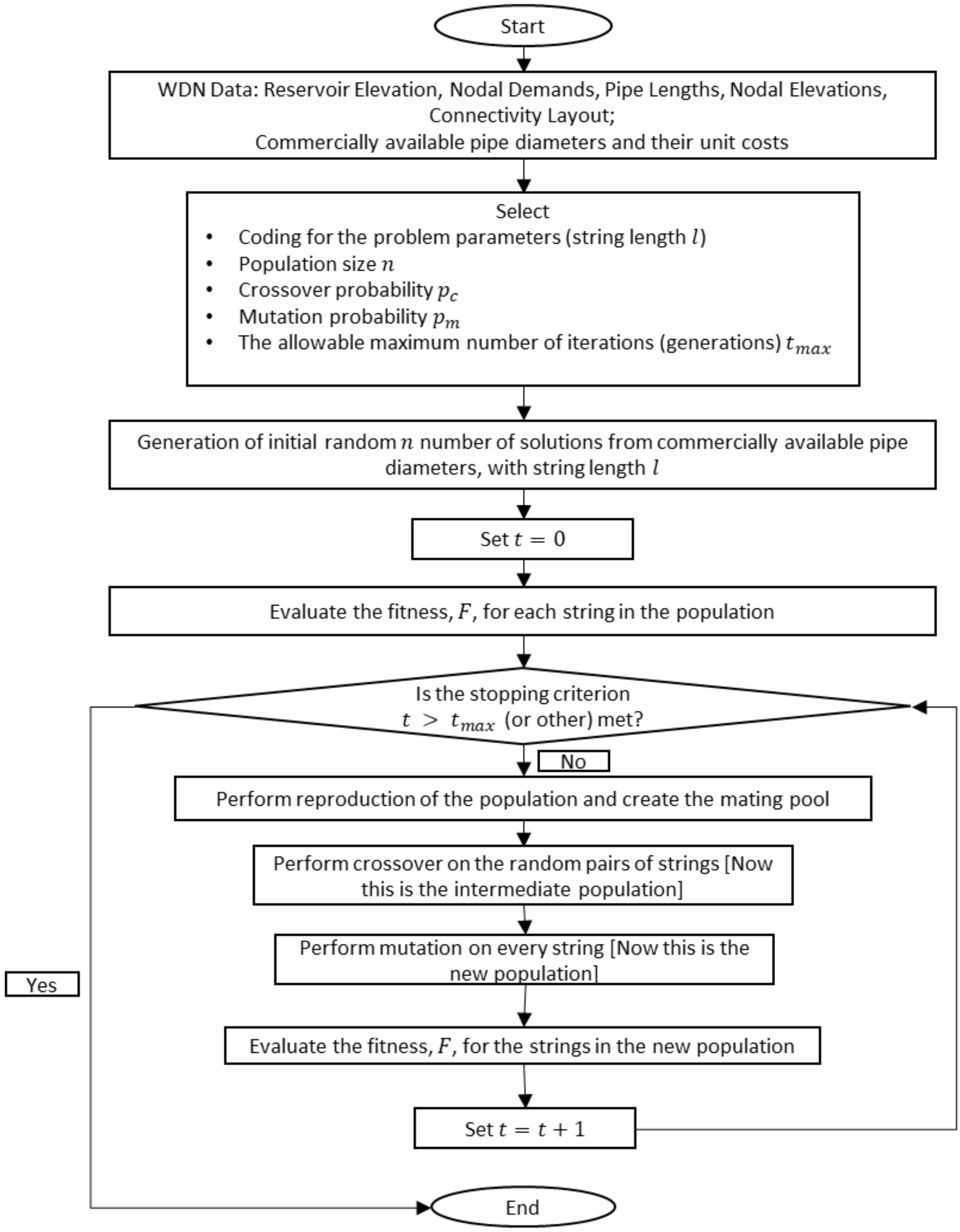

**Figure 3 - Flowchart for the adaptation of the GA considered in this study**

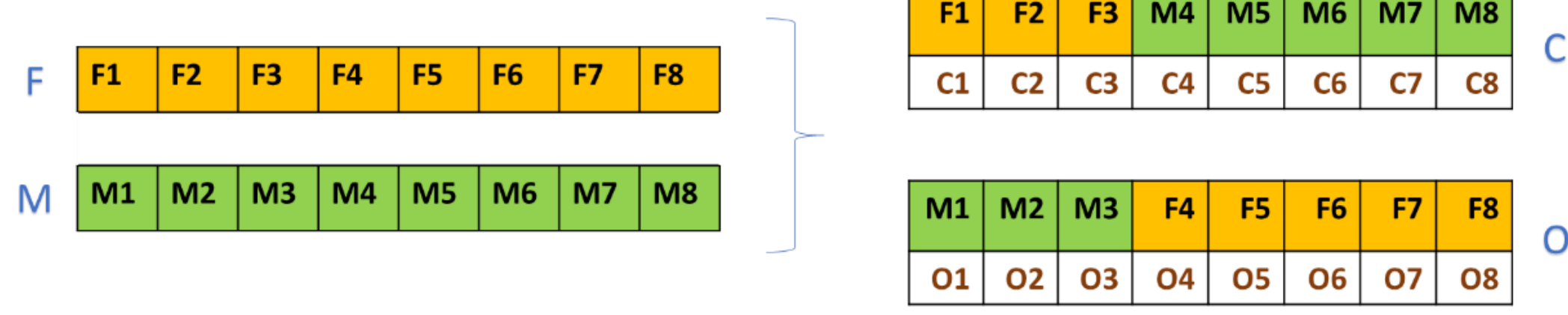

**Figure 4 - The random cut-off point selection and the gene assignment of the crossover operator**

Then, the set of genes of an individual (M), say a "mother", can be represented as follows:

| M1 | M2 | M3 | M4 | M5 | M6 | M7 | M8 |
|---|---|---|---|---|---|---|---|

### 3.4 Fitness Function

The fitness of a particular $k^{th}$ individual, $F_k$, is calculated according to the following formula (7).

$$F_k = \frac{100{,}000}{NP + PP + \sum_{i=1}^{N} C(d_i) * L_i} \qquad (7)$$

where, $NP$ and $PP$ denote nodal penalty, pipe penalty respectively; $C(D_i)$ is the unit cost of the pipe having the diameter $d_i$; $L_i$ is the length of the pipe $i$; and $N$ is the number of pipes involved in the WDN design problem.

The solutions which do not satisfy the minimum hydraulic-head constraint and maximum friction loss constraint, are penalized in proportion to the hydraulic-head deficit $\left(RH_j^{min} - RH_j\right)$ and excess friction loss $(hf_i - hf_{max})$ values, as mentioned in Equation 8 and Equation 9, respectively.

$$NP = Max\left[\left(RH_j^{min} - RH_j\right), 0\right] * (NPF) \qquad (8)$$
$$j = 1,2,3, \dots, nd$$

$$PP = Max[0, (hf_i - hf_{max})] * (PPF) \qquad (9)$$
$$i = 1,2,3, \dots, np$$

### 3.5 Reproduction

If $F_k$ is the fitness of the $k^{th}$ individual, the selection probability of the $k^{th}$ string is proportional to its fitness $F_k$. Note that the population size does not change from generation to generation. The probability for selecting the $k^{th}$ individual for the mating pool is thus calculated as per the formula (10), inspired by the roulette-wheel mechanism.

$$p_k = \frac{F_k}{\sum_{j=1}^{n} F_j} \qquad (10)$$

where $n$ is the total number of individuals in the population.

**Crossover**

Crossover process ensures the mixing of solutions at the global level in the solution space. A single cutoff point is selected at random. Two children are produced from two subsequent parents in the list of mating pool if they are in-line with the *crossover probability*. For one offspring's string, the first parent's genes are placed up to the cutoff point, from which the second parent's genes are placed. For the other offspring's string, similarly, the second parent's genes are placed up to the cutoff point, from which the first parent's genes are placed. The random cut-off point selection is shown in the Figure 4. As demonstrated in the Figure 4, C1, C2, C3, C4, C5, C6, C7 and C8 are the genes of the first child (C) of the parents $F$ and $M$, and O1, O2, O3, O4, O5, O6, O7 and O8 are the genes of the second child (O) of the same parents. The shaded boxes show the corresponding inherited characteristics of parents of these genes.

### 3.6 Mutation

After the crossover operation, the genes of the offspring are changed to any of the other gene values, i. e., any other commercially available pipe diameters, at random. As shown in the Figure 5, the 6th gene of the offspring $C$ is randomly changed from $M6$ to $E6$. This mutation process is carried out to ensure the *local fuzziness* of the solutions obtained by the crossover process which has ensured the *global fuzziness* at this stage. Here, the local fuzziness means the changing of the solution in a mild way, whereas the global fuzziness means the changing of the solution in a drastic way.

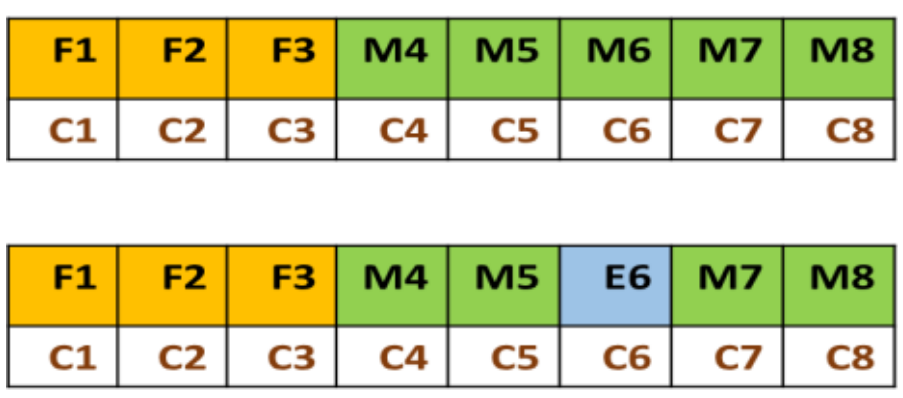

**Figure 5 - Depiction of how mutation of the gene number 6 has occurred**

### 3.7 Gurudeniya Service Zone, Sri Lanka

The general optimization algorithm described under Figure 3, was implemented for the WDN scheme of Gurudeniya Service Zone, Sri Lanka, aiming for obtaining the combination of pipe diameters that minimizes the cost function while satisfying the design constraints. In this study the modern stochastic optimization algorithm, i. e., Genetic Algorithm (GA), was modified as described in the methodology to facilitate the optimal-design of the serial-shaped WDN of Gurudeniya Service Zone. The minimum-allowable-nodal-hydraulic-head value $H_R^{min}$ and the maximum allowable gradient for friction-and-fitting losses $g_{FF}^{max}$ were considered to be 10 m and 0.005 m/m, respectively. The Hazen-William's coefficient, $C_{HW}$, and the fitting loss coefficient, $C_{ft}$, were assumed to be 130 and 1.15, respectively, for all the pipes. Here, the Hazen-William's coefficient is a constant that depends on the material the pipes are made of.

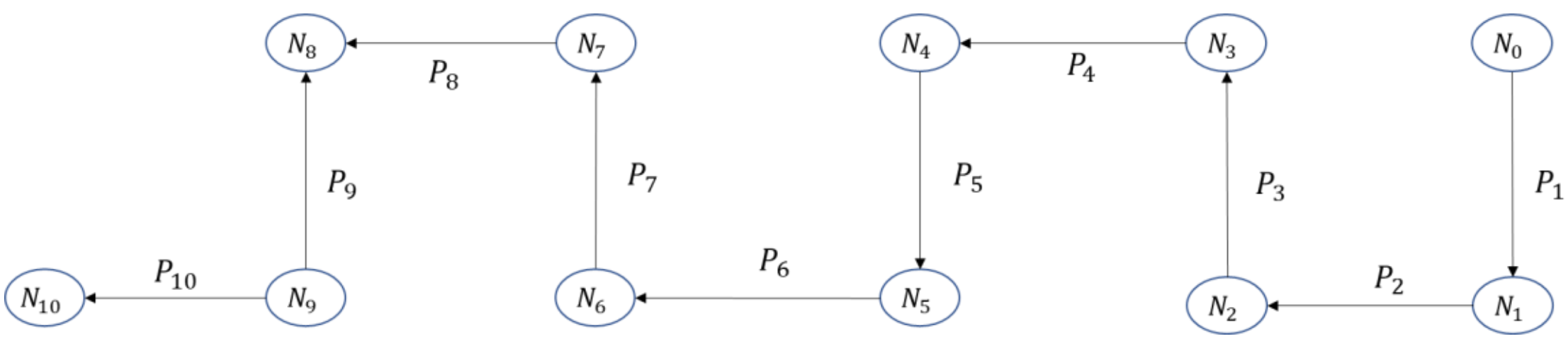

**Figure 6 - Pipe layout directed-graph for WDN of Warapitiya Service Zone, Sri Lanka.**

The data requirement, as identified for the optimization algorithm, included unit-costs of commercially available pipe sizes, reservoir water head, nodal-water-demands, pipe lengths, nodal-elevations, and the pipe-connectivity layout. The pipe layout of Gurudeniya WDN can be seen from the Figure 6 as a directed-graph to identify the flow directions. Pipe length data are given in the Table 1.

**Table 1 - Lengths of pipes being deployed for Gurudeniya WDN (Senavirathna et al., 2022)**

| Pipe ($P_i$) | Length ($L_i$) / ($m$) |
|---|---|
| $P_1$ | 690 |
| $P_2$ | 1120 |
| $P_3$ | 120 |
| $P_4$ | 270 |
| $P_5$ | 630 |
| $P_6$ | 280 |
| $P_7$ | 420 |
| $P_8$ | 230 |
| $P_9$ | 290 |
| $P_{10}$ | 980 |

Nodal-water demand data and nodal-elevation data, including the datum of reservoir elevation, can be referred to as listed in the Table 2. Unit-cost data for different pipe sizes are shown in the Table 3. All these data, were collected from the National Water Supply and Drainage Board (NWS&DB), Sri Lanka.

As the number of pipes involved in this WDN problem, in other words, the number of decision variables of the optimization problem, is equal to 10, the GA string was 10 units in length, having discrete commercially available pipe diameters as its genes. This modified GA was run once selecting a random population of 20 individuals. A high crossover probability of 0.8 was selected in order for the algorithm to facilitate *high-mixing* of solutions. Mutation probability was set to 0.05, a very lower value that is close to zero. Number of generations was assigned to be 5000 for the scope of this study.

**Table 2 - Nodal-water-demands and nodal-elevations of Warapitiya WDN (Senavirathna et al., 2022)**

| Node ($N_j$) | Demand ($D_j$) / ($m^3/day$) | Elevation ($E_j$) / ($m$) |
|---|---|---|
| $N_0$ | -2360.31 | 555 |
| $N_1$ | 796.52 | 452 |
| $N_2$ | 127.50 | 517 |
| $N_3$ | 112.50 | 519 |
| $N_4$ | 165.00 | 535 |
| $N_5$ | 258.76 | 490 |
| $N_6$ | 131.25 | 481 |
| $N_7$ | 168.76 | 476 |
| $N_8$ | 228.76 | 486 |
| $N_9$ | 333.76 | 462 |
| $N_{10}$ | 37.50 | 480 |

**Table 3 - Commercially available pipe sizes and their corresponding unit-costs (Senavirathna et al. [11])**

| Diameter / ($mm$) | Unit-cost |
|---|---|
| 25.40 | 2 |
| 50.80 | 5 |
| 76.20 | 8 |
| 101.60 | 11 |
| 152.40 | 16 |
| 203.20 | 23 |
| 254.00 | 32 |

## 4. Results and Discussion

Table 4 lists the optimal design solution implemented by National Water Supply and Drainage Board, Sri Lanka (NWS&DB), the solution obtained by HBMO algorithm and the solution obtained by GA in this study. The total costs involved with NWS&DB solution, HBMO solution and GA solution are, respectively, 89111, 84520 and 87060. Hence, it is notable that the costs of the global best solutions obtained by HBMO and GA are lower than that of NWS&DB solution, for no violation in the design constraints by indicating zero penalties for all the cases as shown in the Table 4.

**Table 4 - NWS&DB implemented solution, and global best solutions obtained from HBMO (Senavirathna et al. [11]) and GA**

| Pipe $P_i$ | Implemented solution $[d]$ by NWS&DB | Global optimal $[d]$ by HBMO | Global optimal $[d]$ by GA |
|---|---|---|---|
| $P_1$ | 254.0 | 254.0 | 203.2 |
| $P_2$ | 203.2 | 203.2 | 203.2 |
| $P_3$ | 203.2 | 203.2 | 203.2 |
| $P_4$ | 152.4 | 203.2 | 203.2 |
| $P_5$ | 203.2 | 152.4 | 152.4 |
| $P_6$ | 101.6 | 101.6 | 152.4 |
| $P_7$ | 101.6 | 101.6 | 152.4 |
| $P_8$ | 76.2 | 76.2 | 152.4 |
| $P_9$ | 76.2 | 101.6 | 101.6 |
| $P_{10}$ | 76.2 | 50.8 | 50.8 |
| Total Cost | 89,111 | 84,520 | 83,650 |
| Nodal Penalty | 0* | 0 | 0 |
| Pipe Penalty | 0* | 0 | 0 |
| Total Penalty | 0* | 0 | 0 |

- * This is the penalty evaluation for the solution implemented by NWS&DB.

Table 5 and Table 6 give the results, $[g_{FF}]$ and $[H_R]$, respectively, obtained at the hydraulic simulation stage, corresponding to the global optimal solution obtained by GA. It can be seen that from the Table 5, the design constraint $g_{FF\ i} \leq g_{FF\ i}^{max}$ is satisfied for all the pipes ($i = 1,2,\dots,10$) in the serial network by delivering the $g_{FF}$ value of each pipe less than $0.005\ m/m$, whereas the Table 6 shows that the design constraint $H_{R\ j} \geq H_{R\ j}^{min}$ is satisfied for all the nodes ($j = 1,2,\dots,10$) in the serial network by delivering the $H_R$ value of each node greater than $10\ m$. Hence, the application of the global optimal solution obtained by GA for the combination of pipe diameters [d] is appropriate in terms of the satisfaction of the design constraints.

To summarize, the results of the study demonstrate that, GA together with the hydraulic simulation were able to deliver the combination of pipe diameters that minimizes the cost of the given serial pipe configuration network while satisfying the design constraints being concerned.

It can also be seen from Table 4 that, the global optimal solution obtained by GA has confirmed itself to be the best solution when compared with the other solutions obtained by the NWS&DB and HBMO algorithm. It can also intuitively be stated that if the number of generations (at current scope of the study, it is 5000) is increased, the GA's optimal could have been improved in such a way that it can reduce beyond the current global optimal. However, such an investigation may be beyond the scope of this study.

**Table 5 - Pipe-constraints satisfaction demonstration for the solution obtained by GA**

| Pipe $P_i$ | Global optimal $[d]$ by GA | $g_{FF\ i}$ / $(m/m)$ | Design Constraint $g_{FF\ i} \leq g_{FF\ i}^{max}$ satisfied? |
|---|---|---|---|
| $P_1$ | 203.2 | 0.0042 | Yes |
| $P_2$ | 203.2 | 0.0019 | Yes |
| $P_3$ | 203.2 | 0.0016 | Yes |
| $P_4$ | 203.2 | 0.0013 | Yes |
| $P_5$ | 152.4 | 0.0042 | Yes |
| $P_6$ | 152.4 | 0.0025 | Yes |
| $P_7$ | 152.4 | 0.0018 | Yes |
| $P_8$ | 152.4 | 0.0010 | Yes |
| $P_9$ | 101.6 | 0.0024 | Yes |
| $P_{10}$ | 50.8 | 0.0018 | Yes |

**Table 6 - Node-constraints satisfaction demonstration for the solution obtained by GA**

| Node $N_j$ | $H_{R\ j}$ / $(m)$ | Design Constraint $H_{R\ j} \geq H_{R\ j}^{min}$ satisfied? |
|---|---|---|
| $N_1$ | 100.1171 | Yes |
| $N_2$ | 33.0224 | Yes |
| $N_3$ | 30.8328 | Yes |
| $N_4$ | 14.4703 | Yes |
| $N_5$ | 56.8433 | Yes |
| $N_6$ | 65.1481 | Yes |
| $N_7$ | 69.3995 | Yes |
| $N_8$ | 59.1601 | Yes |
| $N_9$ | 82.4587 | Yes |
| $N_{10}$ | 62.6765 | Yes |

## 5. Conclusions

There have been many developments in the computational methods in the field of WDN modeling and optimization, with particular influence from evolutionary computation. Among many criteria for the design of WDNs, one of the widely used is finding the combination of pipe diameters that minimizes the cost of the network while satisfying the minimum-nodal-hydraulic-head and maximum-pipe-head-loss constraints. This paper describes the in-detail GA based combinatorial optimization method used for the optimal design of the water distribution network of Gurudeniya Service Zone, Sri Lanka. Methodology employs fuzzy combinations of pipe diameters to check their suitability to be considered as the cost-effective optimal design solutions. Furthermore, the hydraulic constraints were implicitly evaluated within the GA itself in its aim for reaching the global optimum solution.

It also confirms that the GA based solution obtained in this study is the global optimal solution so far. However, it can be inferred that evolutionary computing such as GA, although cannot guarantee the global optimal solution, can certainly deliver solutions that are in close proximity with the global optimal. Further, this Gurudeniya Service Zone WDN data-set can be used as a benchmark data-set when it comes to testing new computational methods in WDN modeling and optimization.

## Acknowledgements

Authors would like to thank Ms. Imalie Perera from the Department of Mathematical and Statistical Sciences, Clemson University, South Carolina, United States, for sharing her knowledge when implementing the genetic algorithm. Authors also thank National Water Supply and Drainage Board (NWS&DB) for their data used in this study.